\documentclass[conference]{IEEEtran}
\IEEEoverridecommandlockouts
\usepackage{cite}
\usepackage{url}
\newcommand{\method}{DoubleTake}
\usepackage{amsmath,amssymb,amsfonts}
\usepackage{algorithm,algpseudocode}
\usepackage{float,placeins}
\usepackage{booktabs,multirow,tabularx}
\usepackage{graphicx}
\usepackage{textcomp}
\usepackage[table]{xcolor}
\definecolor{oursrow}{HTML}{E4F1ED}
\def\BibTeX{{\rm B\kern-.05em{\sc i\kern-.025em b}\kern-.08em
    T\kern-.1667em\lower.7ex\hbox{E}\kern-.125emX}}
\begin{document}

\title{DoubleTake: Focused Verification of Visual Evidence for Long-Document Understanding}

\author{\IEEEauthorblockN{Kejun Xue, Ang Li, Jinghua Tang, Xincheng Wang, Yongbin Gao, Yanghua Xiao, and Zhijun Fang}}

\maketitle

\begin{abstract}
In long-document question answering, a vision--language model can answer
incorrectly even when the retrieved pages contain sufficient evidence.
Rereading those pages with the original question gives the model little
guidance about which fact to check. We introduce DoubleTake, a training-free
framework for focused verification of visual evidence. An initial reading
produces a candidate answer and a separate evidence record that guides
retrieval for a second reading. A planner compares the two candidates and
their supporting evidence to formulate a factual question. A verifier
answers it from the source images without seeing either candidate or
evidence record, and a reasoner uses this observation to resolve the
original question. With a frozen Qwen3-VL-8B-Instruct backbone, DoubleTake
achieves the highest overall accuracy among the evaluated methods on each
of four benchmarks. With the same pages and number of model calls, asking
a focused factual question yields higher accuracy on all four benchmarks
than asking the original question again.
Our code is available at
\leavevmode\pdfstartlink attr {/Border [0 0 0]} user {/Subtype /Link /A << /S /URI /URI (https://github.com/small-code-cat/double_take) >>}\url{https://github.com/small-code-cat/double_take}\pdfendlink.
\end{abstract}


\section{Introduction}
\label{sec:intro}

In long-document question answering, the evidence needed to answer a
question may be distributed across prose, tables, and charts whose
interpretation depends on their visual
structure~\cite{mathew2021docvqa,masry2022chartqa}.
A vision--language model (VLM) can read page images
directly~\cite{hu2024docowl}, but locating the relevant
information in a long document can require searching across many
pages~\cite{ma2024mmlongbench,deng2025longdocurl}.
Retrieval-augmented generation (RAG) narrows this search to a set of
relevant pages~\cite{faysse2024colpali,cho2024m3docrag}. Within that
set, MDocAgent combines specialized readings of text and images, while
SLEUTH extracts clues and filters pages to build a compact context for
answering~\cite{han2025mdocagent,liu2025sleuth}.

\begin{figure}[!t]
\centering
\includegraphics[width=\columnwidth]{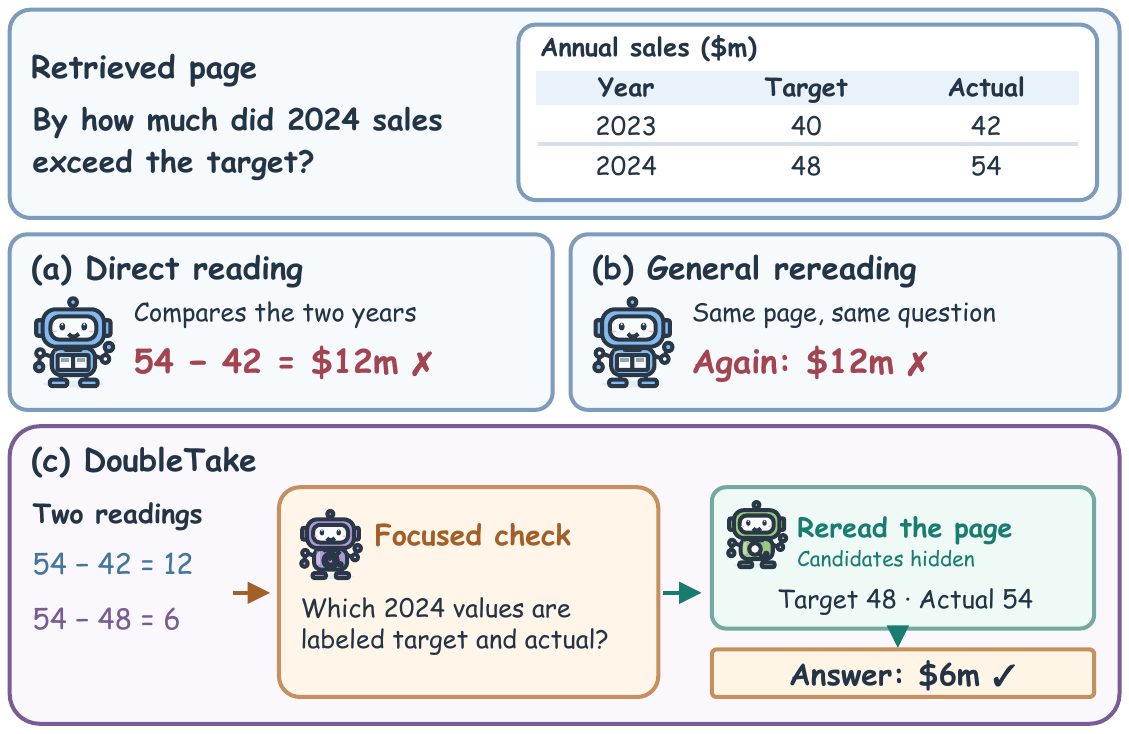}
\caption{Illustrative example. (a) Direct reading compares sales across years. (b) Rereading with the original question repeats the error. (c) DoubleTake uses the two readings to check the 2024 target and actual sales, recovering the correct difference of \$6 million.}
\label{fig:intro-teaser}
\end{figure}

Even when this context contains sufficient evidence, the VLM can still
give an incorrect answer (Fig.~\ref{fig:intro-teaser}(a)). It must
determine which facts are relevant and, when the question calls for a
comparison, how to compare them. This judgment can be checked against the original pages
once a candidate answer is available. Asking the original question
again, however, leaves the VLM to decide for itself which part of its
interpretation needs checking (Fig.~\ref{fig:intro-teaser}(b)). A
focused check therefore requires identifying a fact on which the
proposed answer depends.

Comparing two candidate answers and the evidence behind them can identify
what to check. When the answers differ, their rationales and the evidence
recorded from the pages can reveal which fact would resolve the
disagreement (Fig.~\ref{fig:intro-teaser}(c)). When the answers agree,
the comparison can instead identify a premise shared by both.
In either case, the fact can be expressed as a question that a verifier
can answer directly from the source images without seeing the proposed
answers.

\begin{figure*}[!t]
\centering
\includegraphics[width=\textwidth]{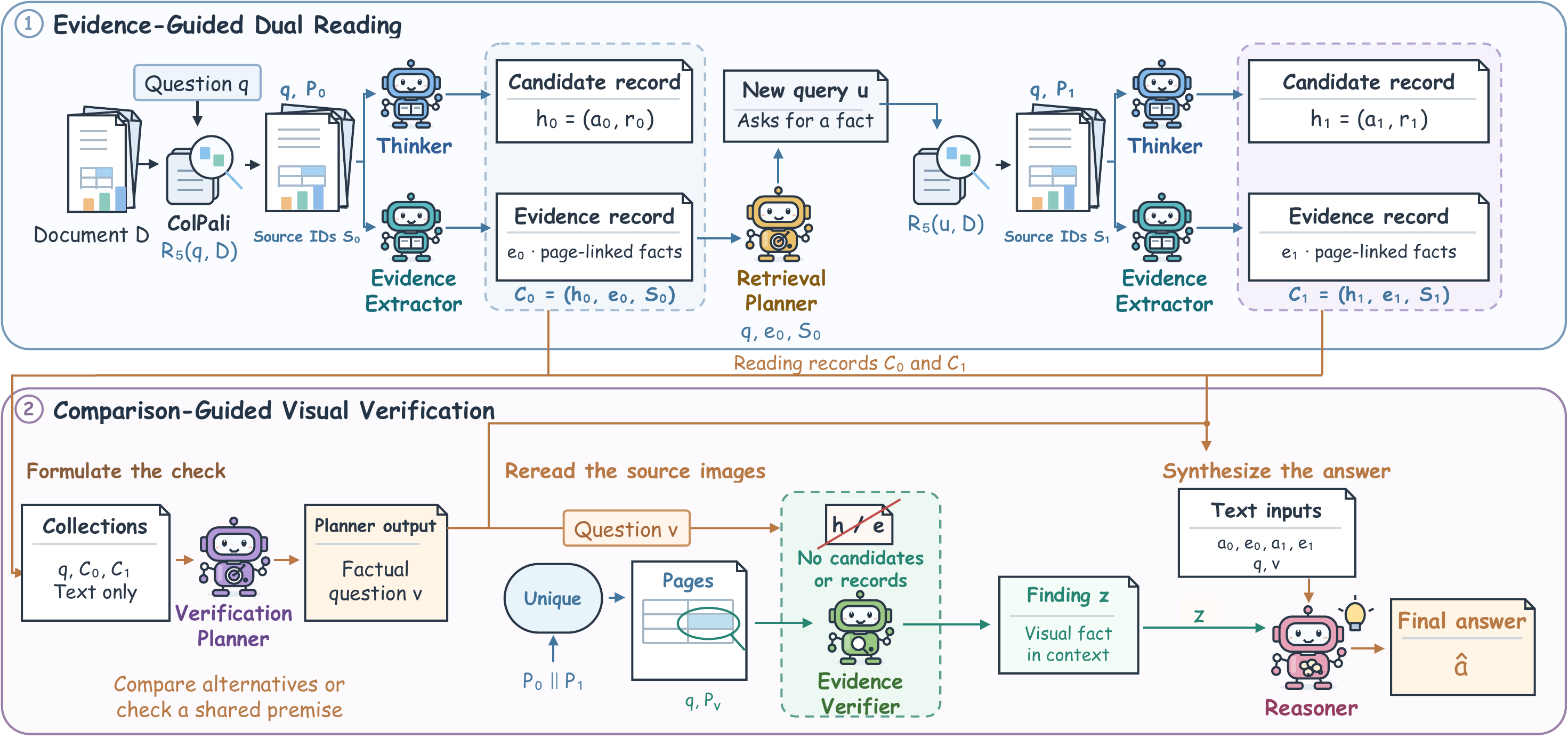}
\caption{DoubleTake connects evidence-guided dual reading with comparison-guided visual verification.
All roles receive the original question. The first evidence record guides the second search. Both candidate--evidence
pairs inform the verification question, and the verifier receives only that
question, the original question, and the retrieved images.}
\label{fig:framework}
\end{figure*}

We introduce \method{}, a training-free framework that uses two
provisional readings to direct visual verification
(Fig.~\ref{fig:framework}). In \emph{evidence-guided dual reading}, the
first reading's evidence record guides a supplementary search. Each
reading produces a candidate answer with its rationale and a separately
extracted record of the page evidence. In \emph{comparison-guided visual
verification}, the Verification Planner compares these readings to
formulate a factual question, and the Evidence Verifier answers it from
the retrieved images without seeing either candidate or evidence record.
The Reasoner combines this observation with both readings to answer the
original question.

Our contributions are summarized as follows:
\begin{itemize}
\item We reveal that a VLM can misidentify which fact applies to the entity or
condition in a question, even when the retrieved pages contain sufficient evidence.
\item We propose \method{}, a training-free framework that compares two
candidate--evidence pairs to select a fact for direct verification against
source images.
\item Across four benchmarks, \method{} achieves the highest overall accuracy
among the evaluated methods, including recent document RAG systems.
\end{itemize}

\section{Related Work}
\label{sec:related}

\textbf{Evidence retrieval and context construction.}
Locating evidence is part of the task in slide decks~\cite{tanaka2023slidevqa},
multi-page documents~\cite{vanlandeghem2023dude}, and reports that span hundreds
of pages~\cite{chia2025mlongdoc}.
In document RAG, ColPali~\cite{faysse2024colpali} matches queries to page images
that preserve table and figure layouts. MDocAgent combines
answers from text and image specialists~\cite{han2025mdocagent}, and SLEUTH
records question-relevant clues from each page while screening images in
parallel. The selected context combines those clues with pages containing
relevant tables, charts, or diagrams for final answering~\cite{liu2025sleuth}.
Feedback from one search can guide the next. Doc-React uses a judge's
feedback to refine its queries~\cite{wu2025docreact}, and ViDoRAG alternates
exploration, summarization, and reflection~\cite{wang2025vidorag}.
HiEvi-RAG verifies which pages support a decomposed subquestion~\cite{xiong2026hievirag},
while VRAG-RL learns when to search and when to zoom in on visual
details~\cite{wang2025vragrl}.
These approaches retrieve, organize, or inspect document evidence to
support answering. A further challenge is to verify how a particular
visual fact relates to the question after provisional answers are formed.

\textbf{Verification and rereading.}
Multiple readings can serve different purposes in question answering.
ReMemR1 revisits earlier memories to recover information during a long
document scan~\cite{shi2026remem}. For visual documents, a further question
is whether a response reflects what the page shows. Zhou et al.\ found that
VLMs relied more heavily on text as the context grew~\cite{zhou2024visualdependency}.
VISA makes an answer's visual support inspectable by marking the relevant
region on a document image~\cite{ma2025visa}.
Self-consistency aggregates answers across sampled reasoning
paths~\cite{wang2023selfconsistency}, while SelfCheckGPT uses disagreement
among sampled responses to flag potentially unsupported
claims~\cite{manakul2023selfcheckgpt}. To check the claims themselves,
CRITIC consults external tools~\cite{gou2024critic}, and
Chain-of-Verification derives factual questions from a draft and answers
them independently before revising it~\cite{dhuliawala2024cove}.
For visual documents, this raises the question of how to formulate a
focused factual check and answer it from the source images.

\section{Methodology}
\label{sec:method}

\subsection{Overall Framework}
\label{sec:overview}

Given a question $q$ and a document $D$ of $N$ indexed page images,
\method{} retrieves pages within $D$ and produces an answer grounded in
those pages. Fig.~\ref{fig:framework} shows how the two designs connect. Evidence-guided
dual reading uses an initial evidence record to retrieve pages for a second
reading. Comparison-guided visual verification uses both readings to select
a fact to check on the retrieved images. The Reasoner receives the resulting
observation together with the candidates and evidence records.

All roles share one frozen vision--language model $\Phi$, with subscripts denoting role-specific prompts. Each call receives only
the inputs listed in Table~\ref{tab:roles}. In particular, the planner sees
both candidates and their rationales when formulating the verification
question. The verifier sees the question and the page images, with both
candidates and both evidence records withheld. Algorithm~\ref{alg:fqu}
gives the order of the two retrievals and eight model calls.

\begin{table}[t]
\centering
\caption{Inputs and outputs by role. For reading $j$, $P_j$ contains page
images, $S_j$ lists their sources, $h_j=(a_j,r_j)$ pairs a candidate answer
with its rationale, and $e_j$ records visual evidence.}
\label{tab:roles}
\footnotesize
\setlength{\tabcolsep}{3pt}
\begin{tabular}{@{}ccc@{}}
\toprule
Role & Input & Output \\
\midrule
Thinker & $q,P_j$ & $h_j$ \\
Evidence Extractor & $q,P_j,S_j$ & $e_j$ \\
Retrieval Planner & $q,e_0,S_0$ & Query $u$ \\
Verification Planner & $q,(h_j,e_j,S_j)_{j=0,1}$ & Question $v$ \\
Evidence Verifier & $q,v,P_v$ & Finding $z$ \\
Reasoner & $a_0,e_0,e_1,a_1,(v,z),q$ & Answer $\hat a$ \\
\bottomrule
\end{tabular}
\end{table}

\subsection{Evidence-Guided Dual Reading}
\label{sec:collections}

\textbf{Initial reading.} For a query $x$, ColPali-v1.3~\cite{faysse2024colpali}
encodes its tokens as vectors $\mathbf t_i(x)$ and each document page $p$
as patch vectors $\mathbf b_k(p)$. Each query token contributes the score of
its best-matching patch. Following ColBERT's late-interaction
scoring rule~\cite{khattab2020colbert}, we sum these contributions:
\begin{equation}
s(x,p)=\sum_i \max_k \langle \mathbf t_i(x),\mathbf b_k(p)\rangle .
\label{eq:page-score}
\end{equation}
The first search retrieves up to five pages in descending score order,
$P_0=R_5(q,D)$.
For either reading $j\in\{0,1\}$, the Thinker and Evidence Extractor process
the same ranked pages $P_j$ in separate calls:
\begin{equation}
\begin{aligned}
h_j=(a_j,r_j)&=\Phi_{\mathrm{think}}(q,P_j),\\
e_j&=\Phi_{\mathrm{extract}}(q,P_j,S_j).
\end{aligned}
\label{eq:reading}
\end{equation}
The Thinker explains its answer $a_j$ in $r_j$. The extractor uses
$S_j$ to link observations to source images, preserving row labels,
headings, and units for table values. Its prompt requests at most
400 words and asks it to flag missing or ambiguous details. For a count,
it lists the visible items. This lets the planner compare the items on
the page with those used in the Thinker's calculation.

\textbf{Supplementary reading.} The first evidence record determines the
next search. The Retrieval Planner formulates a question $u$ about a
missing fact or, when the record appears complete, a premise on which the
answer depends:
\begin{equation}
u=\Phi_{\mathrm{search}}(q,e_0,S_0),\qquad P_1=R_5(u,D).
\label{eq:supp-search}
\end{equation}
The original question supplies the entity, scope, and operation, while
$e_0$ identifies what remains to be found. The Retrieval Planner receives
no initial candidate. For example, if $e_0$ covers only three of the four
quantities needed for a total, $u$ can ask for the fourth by name.
The supplementary search uses $u$ to locate pages, and the extractor and
Thinker independently address the original $q$ on those pages
as in~\eqref{eq:reading}. The two searches can return overlapping pages.

\subsection{Comparison-Guided Visual Verification}
\label{sec:verify}

\textbf{Formulating the check.} The Verification Planner receives each
candidate together with its supporting evidence and page sources. Writing
$C_j=(h_j,e_j,S_j)$ for this collection, the planner produces a single
factual question:
\begin{equation}
v=\Phi_{\mathrm{plan}}(q,C_0,C_1).
\label{eq:check-question}
\end{equation}
The rationales describe the observations and calculations behind each
candidate. The evidence extractor is instructed to record labeled facts
and their source context in a separate call. The planner considers both in light of
the entity and scope of $q$, relating each proposed conclusion to what
its reading reports from the pages. Its prompt asks for a fact that can
distinguish or complete the proposed answers. When the candidates agree, the same
prompt asks about a premise supporting their shared conclusion. The
result is one natural-language question requesting an observation that
can be read directly from the images, such as a labeled value or a
relationship between visual elements.

\textbf{Rereading the pages.} The two ranked lists are concatenated in
retrieval order, retaining the first occurrence of each page. The verifier
receives the resulting images $P_v$, the original question, and the planner's
request:
\begin{equation}
P_v=\operatorname{Unique}(P_0 \mathbin{\|} P_1),
\qquad z=\Phi_{\mathrm{verify}}(q,v,P_v).
\label{eq:verify}
\end{equation}
The question $v$ directs attention to the fact to be checked. The image
set is the ordered union of the two retrievals and contains up to ten
distinct pages. Without access to either candidate or evidence record,
the verifier reports the requested wording, value, or relationship in
$z$ and notes any limits on what it can read.

\textbf{Answer synthesis.} The Reasoner combines the candidate answers
and evidence records with the verification question and finding:

\begin{equation}
\hat a=\Phi_{\mathrm{reason}}\big(a_0,e_0,e_1,a_1,(v,z),q\big).
\label{eq:final-answer}
\end{equation}
Pairing $z$ with its question $v$ preserves the scope of the observation.
The Reasoner interprets this finding alongside the records and applies
the operation requested by $q$. This text-only call can retain a
candidate or combine facts across readings to construct a new answer.
The rationale fields $r_0,r_1$ are used for planning and omitted from
the final message.

\begin{algorithm}[t]
\caption{DoubleTake inference}
\label{alg:fqu}
\small
\begin{algorithmic}[1]
\Require Question $q$, indexed document $D$, retriever $R_5$, frozen model $\Phi$
\State $P_0 \gets R_5(q,D)$;\quad $S_0 \gets \mathrm{Sources}(P_0)$
\State $h_0=(a_0,r_0) \gets \Phi_{\mathrm{think}}(q,P_0)$
\State $e_0 \gets \Phi_{\mathrm{extract}}(q,P_0,S_0)$
\State $u \gets \Phi_{\mathrm{search}}(q,e_0,S_0)$
\State $P_1 \gets R_5(u,D)$;\quad $S_1 \gets \mathrm{Sources}(P_1)$
\State $e_1 \gets \Phi_{\mathrm{extract}}(q,P_1,S_1)$
\State $h_1=(a_1,r_1) \gets \Phi_{\mathrm{think}}(q,P_1)$
\State $v \gets \Phi_{\mathrm{plan}}(q,(h_0,e_0,S_0),(h_1,e_1,S_1))$
\State $P_v \gets \mathrm{Unique}(P_0 \mathbin{\|} P_1)$
\State $z \gets \Phi_{\mathrm{verify}}(q,v,P_v)$
\State \Return $\Phi_{\mathrm{reason}}(a_0,e_0,e_1,a_1,(v,z),q)$
\end{algorithmic}
\end{algorithm}

\FloatBarrier
\section{Experiments}
\label{sec:experiments}

\subsection{Experimental Setup}
\label{sec:setup}

\begin{table*}[!t]
\centering
\caption{Question-level accuracy (\%) across four benchmarks. Bold and underlining mark the highest and second-highest distinct scores in each column, including ties.}
\label{tab:main}
\footnotesize
\setlength{\tabcolsep}{2.4pt}
\begin{tabularx}{\textwidth}{>{\centering\arraybackslash}X*{14}{c}}
\toprule
\multirow{2}{*}{Method} & \multicolumn{7}{c}{MMLongBench-Doc} &
\multicolumn{4}{c}{LongDocURL} & \multirow{2}{*}{PaperTab} &
\multirow{2}{*}{FetaTab} & \multirow{2}{*}{Micro-avg.} \\
\cmidrule(lr){2-8}\cmidrule(lr){9-12}
 & Chart & Table & Pure-text & Layout & Figure & None & Overall
 & Locating & Understanding & Reasoning & Overall & & & \\
\midrule
Base & 36.52 & 35.32 & 42.30 & 38.66 & 38.49 & \textbf{81.12} & 50.09 & 44.46 & 74.90 & 57.11 & 62.84 & 38.93 & 72.83 & 60.13 \\
M3DocRAG & 41.01 & 37.16 & 44.92 & 42.86 & 44.74 & 40.77 & 44.55 & \textbf{47.63} & 77.39 & 56.59 & 65.03 & 48.85 & 75.20 & 61.25 \\
MoLoRAG & \underline{50.00} & 42.66 & 48.20 & 37.82 & 40.79 & 56.65 & 49.26 & 38.71 & 65.57 & 58.66 & 56.39 & 48.09 & 74.70 & 57.97 \\
SLEUTH & 38.20 & 36.70 & 42.30 & 37.82 & 45.39 & 66.09 & 49.35 & 41.87 & 71.28 & 56.85 & 60.09 & 37.15 & 71.06 & 58.12 \\
MDocAgent & \textbf{54.49} & \textbf{53.21} & \underline{56.07} & \underline{48.74} & \underline{50.99} & 53.22 & \underline{54.62} & 43.74 & \textbf{79.57} & \underline{65.63} & \underline{66.54} & \underline{54.71} & \underline{77.26} & \underline{65.16} \\
\rowcolor{oursrow}
\method{}~(Ours) & \textbf{54.49} & \underline{51.38} & \textbf{57.38} & \textbf{51.26} & \textbf{52.96} & \underline{70.39} & \textbf{59.33} & \underline{46.04} & \underline{77.55} & \textbf{70.28} & \textbf{66.92} & \textbf{56.23} & \textbf{78.35} & \textbf{66.76} \\
\bottomrule
\end{tabularx}
\end{table*}

\begin{table*}[!t]
\centering
\caption{Accuracy (\%) with no verification (NV), no supplementary reading (NS), no Thinker calls (NT), or general rereading (GR). Bold and underlining mark the highest and second-highest distinct scores in each column, including ties.}
\label{tab:ablation}
\footnotesize
\setlength{\tabcolsep}{2.4pt}
\begin{tabularx}{\textwidth}{>{\centering\arraybackslash}X*{14}{c}}
\toprule
\multirow{2}{*}{Variant} & \multicolumn{7}{c}{MMLongBench-Doc} &
\multicolumn{4}{c}{LongDocURL} & \multirow{2}{*}{PaperTab} &
\multirow{2}{*}{FetaTab} & \multirow{2}{*}{Micro-avg.} \\
\cmidrule(lr){2-8}\cmidrule(lr){9-12}
 & Chart & Table & Pure-text & Layout & Figure & None & Overall
 & Locating & Understanding & Reasoning & Overall & & & \\
\midrule
\rowcolor{oursrow}
\method{}~(Ours) & \underline{54.49} & \textbf{51.38} & 57.38 & \textbf{51.26} & \textbf{52.96} & 70.39 & \textbf{59.33} & 46.04 & \underline{77.55} & \textbf{70.28} & \textbf{66.92} & \underline{56.23} & \textbf{78.35} & \textbf{66.76} \\
NV & 53.93 & \underline{50.00} & 56.07 & 47.06 & 47.04 & \textbf{73.82} & 57.86 & 42.88 & 75.38 & 65.63 & 64.04 & \underline{56.23} & 76.08 & 64.56 \\
NS & 53.37 & 46.33 & 53.77 & \textbf{51.26} & 49.67 & 72.10 & 56.93 & \textbf{46.76} & 77.15 & 66.15 & 66.24 & 52.16 & 76.87 & 65.24 \\
NT & \textbf{55.62} & \textbf{51.38} & \underline{57.70} & \underline{49.58} & 49.67 & \underline{72.96} & \underline{59.24} & 45.18 & 77.39 & 67.44 & 66.11 & \textbf{57.25} & 76.87 & \underline{66.11} \\
GR & 53.37 & \underline{50.00} & \textbf{58.03} & 47.06 & \underline{51.32} & 69.10 & 57.67 & \underline{46.19} & \textbf{77.63} & \underline{69.25} & \underline{66.84} & 54.96 & \underline{77.07} & 65.97 \\
\bottomrule
\end{tabularx}
\end{table*}

\textbf{Datasets and Evaluation Metrics.} We evaluate on four document QA
benchmarks. LongDocURL~\cite{deng2025longdocurl} tests whether a model can
locate, extract, and reason over information in long documents (2,325 questions).
PaperTab and FetaTab~\cite{hui2024uda} ask about tables and their surrounding
text in research papers (393 questions) and Wikipedia articles (1,016 questions),
respectively. MMLongBench-Doc (MMLB)~\cite{ma2024mmlongbench} includes
1,082 questions spanning text, tables, charts, figures, and layout, including
questions the document cannot answer. MMLB evidence categories can overlap.
None denotes an empty evidence-source list. LongDocURL assigns one task
category to each question.
We report accuracy on each benchmark. Micro-avg. is the percentage of
all 4,816 questions answered correctly, giving each question equal weight. For every method, \texttt{qwen3.8-flash} judges whether the
prediction answers the question correctly given the reference, with thinking
disabled and temperature set to zero.

\textbf{Baselines.} All methods use the same questions and judge.
(1) \emph{Base} answers the original question directly from the retrieved pages.
(2) \emph{M3DocRAG}~\cite{cho2024m3docrag} uses the top five ColPali page
images with a short-answer prompt in our implementation.
(3) \emph{MoLoRAG}~\cite{wu2025molorag} selects pages by beam search over
a page graph using semantic and logical relevance scores.
(4) \emph{MDocAgent}~\cite{han2025mdocagent} combines answers from text and
image experts, using BGE-M3~\cite{chen2024m3embedding} for text retrieval and Qwen3-8B~\cite{qwen3technicalreport} for its text branch.
(5) \emph{SLEUTH}~\cite{liu2025sleuth} extracts clues from retrieved pages,
filters the pages, and answers from the resulting context.

\textbf{Implementation Details.} By default, all roles share a frozen
Qwen3-VL-8B-Instruct~\cite{bai2025qwen3vl}. ColPali-v1.3~\cite{faysse2024colpali}
retrieves up to five pages from the given document for each query. We distribute questions
across eight RTX A6000 GPUs and run inference in BF16. We use a temperature
of 0.1, top-$p$ of 0.8, top-$k$ of 20, a repetition penalty of 1.1, and a
limit of 8,192 new tokens per call.

\subsection{Overall Performance}
\label{sec:main}

\method{} achieves the highest overall accuracy on all four benchmarks
(Table~\ref{tab:main}). Its largest lead over MDocAgent is on MMLB,
where questions with no annotated evidence source account for 40 of the
51 additional correct answers. Among annotated categories, \method{} leads
on layout and figures, ties on charts, and trails MDocAgent on tables.
On LongDocURL, \method{} leads MDocAgent by 4.65 percentage points
on reasoning questions.
MDocAgent leads on understanding and has higher overall accuracy on the
subset for which both methods produced an answer.

\subsection{Quantitative Analysis}
\label{sec:verif-analysis}

\subsubsection{Component Ablations}
Table~\ref{tab:ablation} shows the effect of removing each component.
Without verification (NV), the Reasoner still receives both readings,
but micro-averaged accuracy falls more than in the other component
ablations. The drop is especially clear on MMLB figure questions, where
accuracy falls by nearly six points. Reopening the image provides an
observation for assessing the earlier interpretation.
Removing the supplementary reading (NS) reduces accuracy on every
benchmark, with the largest drop on PaperTab, from 56.23\% to 52.16\%.
This ablation removes the supplementary pages, record, and candidate
together. With both readings retained, the variant without candidate
generation (NT) reaches 57.25\% on PaperTab, showing that the planner
and Reasoner can also use the evidence records without candidate answers.

\begin{table}[!t]
\centering
\caption{Final answer accuracy for completed questions, grouped by initial and supplementary candidate correctness ($\checkmark$ correct, $\times$ incorrect). Overall pools the four benchmarks.}
\label{tab:candidate-correctness}
\footnotesize
\setlength{\tabcolsep}{1.5pt}
\begin{tabular*}{\columnwidth}{@{\extracolsep{\fill}}*{7}{c}@{}}
\toprule
\multicolumn{2}{c}{Candidate} & \multicolumn{5}{c}{Final accuracy (\%)} \\
\cmidrule(lr){1-2}\cmidrule(lr){3-7}
Initial & Suppl. & MMLB & LongDocURL & PaperTab & FetaTab & Overall \\
\midrule
$\times$     & $\times$     & 13.3 & 11.2 & 15.6 & 17.8 & 13.2 \\
$\checkmark$ & $\times$     & 63.4 & 60.7 & 73.5 & 65.6 & 63.7 \\
$\times$     & $\checkmark$ & 72.5 & 71.3 & 60.4 & 67.3 & 69.0 \\
$\checkmark$ & $\checkmark$ & 94.1 & 95.7 & 94.0 & 97.6 & 95.8 \\
\bottomrule
\end{tabular*}
\end{table}

\subsubsection{Focused versus General Rereading}
General rereading (GR) uses the same candidates, records, images, and
eight calls as \method{}, including the planner call. Its reader revisits
the original question. In both settings, the Reasoner receives the
question posed to the reader and the resulting observation. The focused
request yields higher accuracy on all four benchmarks. The difference is
clearest on MMLB layout questions (51.26\% versus 47.06\%), where GR
matches the accuracy of skipping verification. A focused request can
ask the verifier to identify the heading associated with a value.

\subsubsection{Sensitivity to Retrieval Depth}
\begin{figure}[!t]
\centering
\includegraphics[width=\columnwidth]{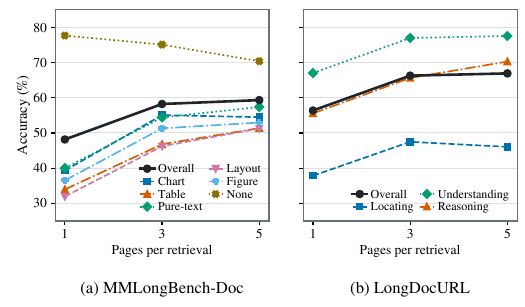}
\caption{DoubleTake accuracy by retrieval depth and category. Both searches use the
same page limit. MMLB evidence categories may overlap; None denotes an
empty evidence-source list.}
\label{fig:retrieval-depth}
\end{figure}

Increasing the limit from one to three pages per search improves overall
accuracy by about ten points on both benchmarks (Fig.~\ref{fig:retrieval-depth}).
On MMLB, all five annotated evidence categories improve from one to five
pages, while None falls from 77.68\% to 70.39\%. On LongDocURL, increasing
the limit from three to five raises reasoning accuracy from 65.63\% to
70.28\%, while locating falls from 47.48\% to 46.04\%.

\subsubsection{Performance Across Backbones}
\begin{figure}[!t]
\centering
\includegraphics[width=\columnwidth]{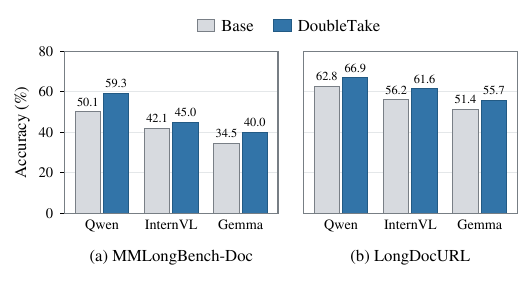}
\caption{Final answer accuracy with Qwen3-VL-8B-Instruct, InternVL3.5-8B~\cite{wang2025internvl3_5}, and Gemma-3-12B-IT~\cite{gemma_2025}. Base answers directly from the initial retrieval,
and \method{} uses the same backbone throughout its full pipeline.}
\label{fig:backbones}
\end{figure}

Across Qwen3-VL-8B-Instruct, InternVL3.5-8B~\cite{wang2025internvl3_5}, and Gemma-3-12B-IT~\cite{gemma_2025}, \method{} outperforms direct answering on both
benchmarks (Fig.~\ref{fig:backbones}). With Gemma on LongDocURL, reasoning
questions account for 55 of the 100 net additional correct answers,
although they make up only one sixth of the benchmark. On MMLB, all three backbones gain on the five annotated evidence types but lose accuracy on questions without an annotated source.

\subsection{Final Accuracy by Candidate Correctness}
\label{sec:candidate-correctness}
When exactly one candidate is correct, \method{} answers 391 of 591
questions correctly (66.2\%; Table~\ref{tab:candidate-correctness}).
Always choosing the initial or supplementary candidate would yield
53.6\% or 46.4\%, respectively. When both candidates are wrong, the
final answer is correct in 13.2\% of completed questions in that group.

\begin{figure}[!t]
\centering
\includegraphics[width=\columnwidth]{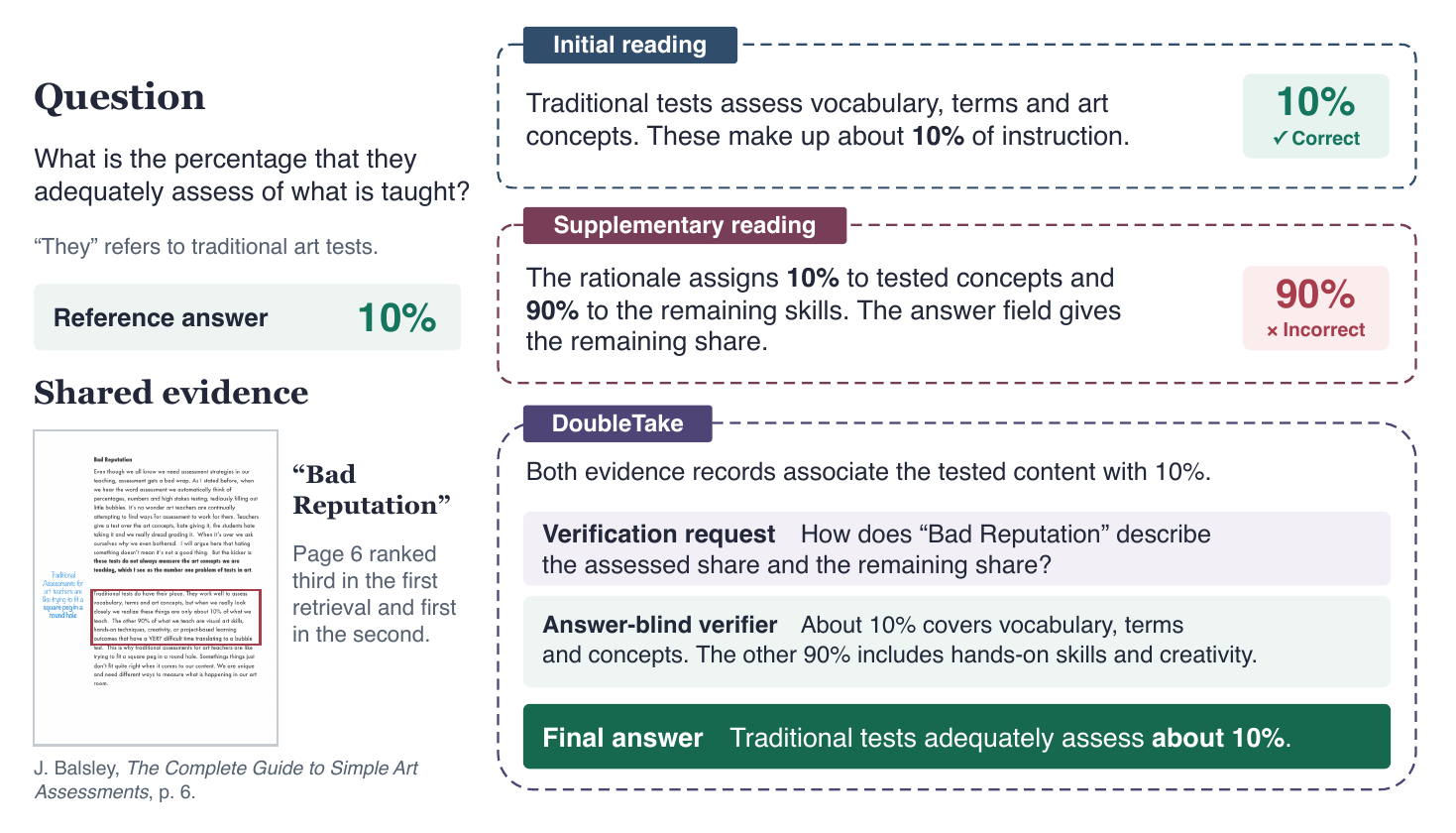}
\caption{Two readings of the same page yield different answers to
LongDocURL question 1498. The red box marks the passage in
J.~Balsley's \emph{The Complete Guide to Simple Art Assessments}, p.~6.}
\label{fig:case1498}
\end{figure}

\subsection{Case Study}
\label{sec:cases}
In Fig.~\ref{fig:case1498}, both readings retrieve the evidence page,
yet the supplementary reader reports the 90\% that traditional tests
struggle to assess. Both evidence records associate 10\% with assessed content, even though the candidates disagree. Asked what each percentage measures, the verifier
identifies 10\% as the assessed share, and the Reasoner retains the
initial answer.

\FloatBarrier
\section{Conclusion}
\label{sec:conclusion}

\method{} compares two readings of a document to identify a fact on which
the answer depends and checks it against the source images. The resulting
observation can support a candidate answer or supply a missing fact needed
to construct a different answer. \method{} achieved the highest overall
accuracy among the evaluated methods on each of four benchmarks. On all
four, asking the verifier a focused factual question yielded higher final
answer accuracy than asking it the original question, with the same pages
and the same number of model calls. Future work will examine how to use
the two readings to decide when to initiate verification.


\clearpage
\IEEEtriggeratref{16}
\bibliographystyle{IEEEtran}
\bibliography{references}

\end{document}